%% file: acl2021.tex
\pdfoutput=1
\documentclass[11pt]{article}

\usepackage{hyperref}
\usepackage{multirow}
\usepackage{booktabs}
\usepackage{tabularx}
\usepackage{stfloats}

\usepackage[]{ACL2023}

\usepackage{times}
\usepackage{latexsym}
\usepackage{booktabs}
\usepackage{xspace}
\usepackage{color,soul}
\usepackage{balance}
\usepackage{paralist}
\usepackage{verbatim}
\usepackage[normalem]{ulem}
\usepackage{amsmath}

\usepackage[T1]{fontenc}

\usepackage[utf8]{inputenc}
\usepackage{txfonts}
\usepackage{microtype}

\usepackage{subcaption}
\usepackage{graphicx}
\usepackage{makecell}

\definecolor{cInfusion}{HTML}{0367A6}
\definecolor{cGREEN}{HTML}{4c9900}
\definecolor{cYELLOW}{HTML}{FDBC42}
\definecolor{cBLUE}{HTML}{0071BC}
\definecolor{cRED}{HTML}{ED1B23}

\definecolor{cpre}{HTML}{6EB4FD}
\definecolor{clabel}{HTML}{F89151}
\definecolor{cexp}{HTML}{99C893}

\newcommand{\metricname}{\textcolor{cGREEN}{ \textsc{Treu\xspace} }}

\newcommand{\Infusion}{\texttt{Infusion\xspace} }
\newcommand{\Baseline}{\texttt{Baseline\xspace} }

\title{Are Human Explanations Always Helpful? \\Towards Objective Evaluation of Human Natural Language Explanations}

\author{ Bingsheng Yao  \\ Rensselaer Polytechnic Institute \\
\And Prithviraj Sen \thanks{$^{\dagger}$ Work done while Prithviraj was at IBM Research.}\\ Amazon \\
\And Lucian Popa \\ IBM Research \\
\AND James Hendler \\ Rensselaer Polytechnic Institute \\
\And Dakuo Wang \thanks{ $^{\dagger}$\texttt{d.wang@northeastern.edu} Corresponding Author. }\\ Northeastern University \\
}

\begin{document}
\maketitle
\begin{abstract}


Human-annotated \textbf{labels} and \textbf{explanations} are critical for training explainable NLP models. 
However, unlike human-annotated \textbf{labels} 
whose quality is easier to calibrate 
(e.g., with a majority vote),  human-crafted \textbf{free-form explanations} can be quite subjective.
Before blindly using them as ground truth to train ML models, a vital question needs to be asked: \textbf{How do we evaluate a human-annotated explanation’s quality?}
In this paper, we 
build on the view 
that the quality of a human-annotated explanation 
can be measured based 
on its helpfulness (or impairment) to the ML models’ performance for the desired NLP tasks for which the annotations were collected.
In comparison to the commonly used \texttt{Simulatability} score, we define a new metric that can take into consideration of the helpfulness of an explanation for model performance at both fine-tuning and inference.  
With the help of a unified dataset format, we evaluated the proposed metric on five datasets (e.g., e-SNLI) against two model architectures (T5 and BART), and the results 
show 
that our proposed metric can objectively evaluate the quality of human-annotated explanations, while \texttt{Simulatability} falls short.  

\end{abstract}

\input{sections/intro}
\input{sections/relate}

\input{sections/Unified_Structure}

\input{sections/Preliminary_Experiment}

\input{sections/eval}

\input{sections/conclusion}

\input{sections/limitations}

\input{sections/ethics}

\input{sections/ack}

\newpage
\bibliography{anthology, custom}
\bibliographystyle{acl_natbib}

\clearpage
\appendix

\input{sections/appendix}

\end{document}

%% file: sections/intro.tex
\section{Introduction}



Despite the recent advances of large-scale language models (LLM)~\citep{devlin-etal-2019-bert, qin2023chatgpt, lewis2019bart, raffel2020exploring}, which exhibit close-to-human performance on many natural language processing (NLP) tasks (e.g., Question Answering~\citep{rajpurkar2016squad, kovcisky2018narrativeqa, mou-etal-2020-frustratingly, mou-etal-2021-narrative, xu-etal-2022-fantastic}, Natural Language Inference~\citep{bowman2015large, williams2017broad, wang2018glue}, and Text Generation~\citep{duan2017question, yao-etal-2022-ais,zhao2022educational}), humans are  eager to know how State-of-the-Art (SOTA) models arrive at a prediction. 
Researchers working on natural language explanations\footnote{In this paper, we use ``explanations'' and ``natural language explanations'' to refer to the collective concepts of ``free-text rationales'' and ``natural language explanation'', which differ from ``rule-based'' or ``extractive'' explanations.} turned to human annotators for help by recruiting crowd-workers or  experts to annotate both the labels and corresponding natural language explanations~\citep{camburu2018snli, rajani2019explain, aggarwal2021explanations,wang2019designing}; 
Researchers can thus leverage human-annotated explanations to boost models' prediction performance or train models to generate human-understandable natural language explanations. 

However, the quality issue of human-annotated explanations has yet to be explored.  
Researchers often 
leverage popular Natural Language Generation (NLG) metrics such as \texttt{BLEU}~\citep{papineni2002bleu} and \texttt{ROUGE}~\citep{lin2004rouge} 
to evaluate the similarity between model-generated and human-annotated explanations, with a strong assumption that human-annotated ones are the gold standard. 
Nevertheless, 
unlike providing labels for classification or multiple-choice QA tasks~\cite{chen2021goldilocks}, different people may come up with distinct 
natural language explanations for the same observation~\cite{gebreegziabher2023patat}. Two such explanations can be both correct even though the BLEU or ROUGE similarity may be low. Furthermore, human-given natural language explanations can often be subjective and 
task-dependent~\cite{lee2022evaluating}. 
As a result, human-annotated explanations should not be simply treated as the gold standard~\cite{muller2021designing};
instead, we take the view that 
the core value of explanations should be based on how much help they provide towards the model prediction instead of being based on notions of semantic similarity or word-matching. 

\begin{table*}[!ht]
\centering
\resizebox{.98\textwidth}{!}{%

\begin{tabular}{lcccccc} 

\toprule

\multirow{2}{*}{ Dataset }       &    \multirow{2}{*}{ Task }        &   \multirow{2}{*}{ \begin{tabular}[c]{@{}c@{}} Task Format \end{tabular} }    &    \multicolumn{3}{c}{ Data Instances }    &  \multirow{2}{*}{ \begin{tabular}[c]{@{}c@{}} Average explanation \\ Length  (token) \end{tabular} } \\
\cmidrule{4-6}
    &   &   &   Train   &   Valid   &   Test    &   \\
    
\cmidrule(lr){1-1} \cmidrule(lr){2-2} \cmidrule(lr){3-3} \cmidrule(lr){4-4} \cmidrule(lr){5-5} \cmidrule(lr){6-6} \cmidrule(lr){7-7} 

CoS-E v1.0      &   Commonsense QA              & 3-choice Multiple-Choice  & 7610      &   950     &   -       &   16.148 \\ 
CoS-E v1.11     &   Commonsense QA              & 5-choice Multiple-Choice  & 9741      &   1221    &   -       &   8.996 \\
ECQA            &   Commonsense QA              & 5-choice Multiple-Choice  & 7598      &   1098    &   2194    &   63.572 \\
e-SNLI          &   Natural Language Inference  & 3-label Classification    & 549367    &   9842    &   9824    &   15.977 \\
ComVE           &   Commonsense Validation      & 2-choice Multiple-Choice  & 10000     &   1000    &   1000    &   10.288 \\

\bottomrule
\end{tabular}
}

\vspace{-0.5em}
\caption { Task description and core statistics for five popular large-scale datasets with human-annotated natural language explanations that are included in our evaluation.  }
\vspace{-1em}
\label{tab:dataset_stats}
\end{table*}

To summarize our contributions in this paper: \\
\indent 1. 
We provide an objective evaluation to quantify the human-annotated explanations' 
helpfulness towards model performance. 
Our evaluation metric is an extension of the \texttt{Simulatability} score~\citep{doshi2017towards} 
and we propose a prompt-based unified data format that can convert classification or multiple choice tasks into a unified multiple choice generation task format 
to minimize the influence of structural variations across different tasks.\\
\indent 2.
Through an evaluation with five datasets and two models, our metric can rank explanations quality consistently across all five datasets on two model architectures while the \texttt{Simulatability} score (baseline) falls short.\\
\indent 3. 
Our evaluation justifies the hypothesis that human explanations can still benefit model prediction, even if they were criticized as low-quality by prior literature's human evaluation. \\

%% file: sections/relate.tex
\vspace{-1em}
\section{Related Work}

\subsection{Natural Language Explanation Datasets}


Despite the development of new model architectures and potentially more significant parameters, these ``black boxes'' unavoidably lack the ability to explain their predictions; this led to increased efforts in the community to leverage 
human-annotated explanations to either train models with explanations or to teach them to self-rationalize. For example, \citet{wiegreffe2021teach} reviewed 65 datasets and provided a 3-class taxonomy of explanations: highlights, free-text, and structured. We focus on five large public datasets with free-text human-annotated explanations at the instance level (Table~\ref{tab:dataset_stats}). We double-checked these datasets' licenses, and no personally identifiable information (PII) exists. 

One prominent dataset is CoS-E and its two variants \textbf{CoS-E v1.0} and \textbf{CoS-E v1.11}\citep{rajani2019explain}.
It extended the Commonsense Question-Answering (CQA v1.0 and v1.11 versions) dataset \citep{talmor2018commonsenseqa} by adding human-annotated explanations to the correct answer label. 
However, a few recent works suggest that the CoS-E's explanation quality is not good, as \citet{narang2020wt5} independently hand-labeled some new explanations for CoS-E and found a very low BLEU score between its original explanations and the new ones.
To improve the explanation's quality, \textbf{ECQA} \citep{aggarwal2021explanations} collected and summarized single-sentence explanation for each candidate answer into a natural language explanations for every data in the CQA v1.11 dataset.
\citet{sun2022investigating} proved that CoS-E explanations are not as good as ECQA explanations based on human preferences. 
The fourth dataset is \textbf{e-SNLI}\citep{camburu2018snli}, which consists of explanations for the Stanford Natural Language (SNLI) dataset \citep{bowman2015large}. Finally, the fifth dataset is \textbf{ComVE} \citep{wang2020semeval}, asking which one of two sentences is against commonsense. 
Later we evaluate the human-annotated explanations in the above-mentioned five datasets with our metric and an established baseline, the \texttt{Simulatability} score.

Worth mentioning that we do not include datasets such as \textbf{SBIC} \citep{sap2019social} or\textbf{ E-$\delta$-NLI }\citep{brahman2021learning}. SBIC does not provide explanations for all the data, and E-$\delta$-NLI leverages various sources to augment the $\delta$-NLI \citep{rudinger2020thinking} dataset with explanations instead of providing human annotations.

\begin{figure*}[!tp]
    \centering
    \includegraphics[width=.98\textwidth]{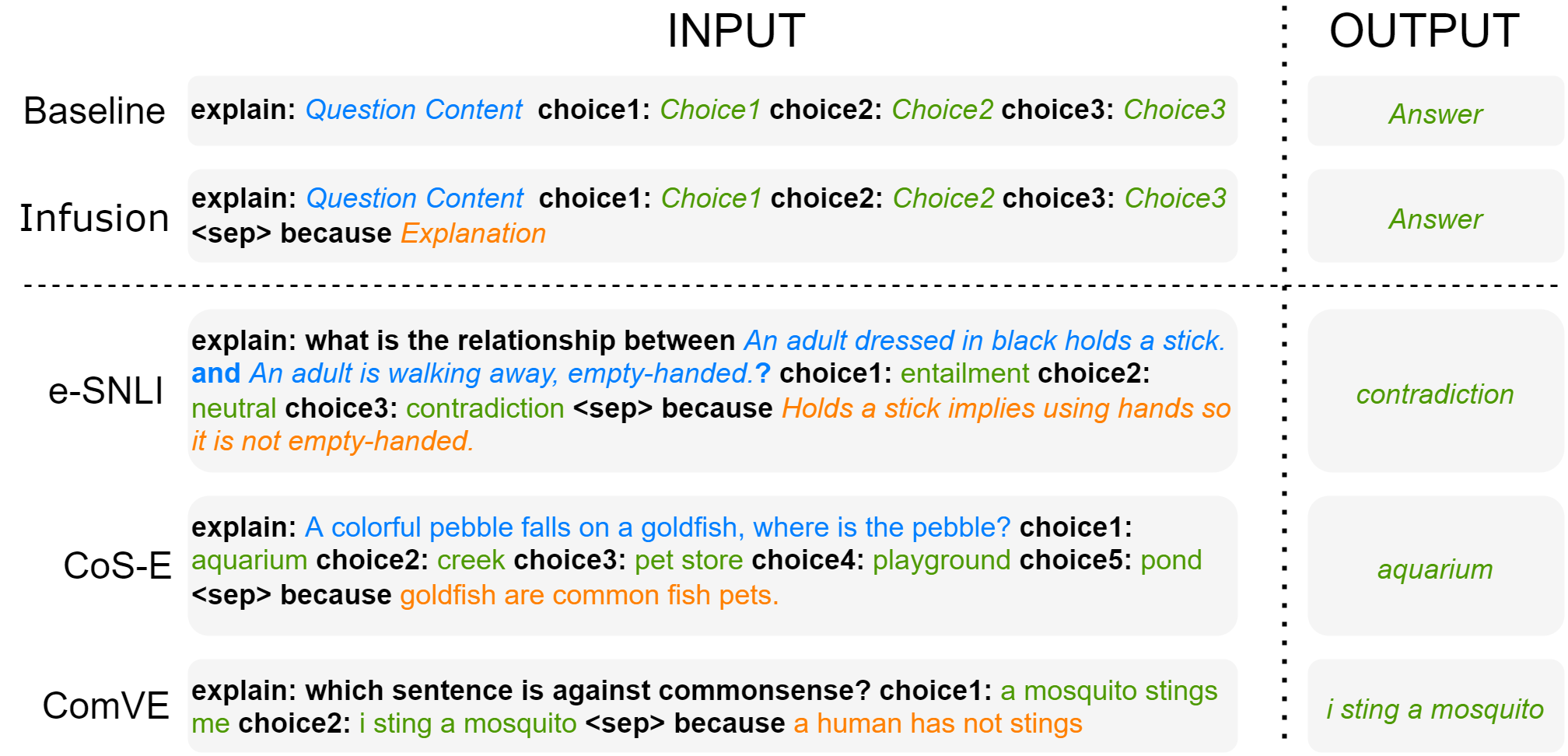}
    \vspace{-0.5em}
    \caption{ Unified structure of \Baseline and \Infusion settings. Black bold text are fixed prompts. We provide examples of \Infusion format in classification task like e-SNLI and multiple choice task like CoS-E and ComVE. The color schema follows: \textcolor{cpre}{blue} denotes question content; \textcolor{cexp}{green} denotes choice content; \textcolor{clabel}{orange} denotes explanations. }
    \vspace{-1em}
    \label{fig:unified_format}
\end{figure*}

\subsection{Evaluation Metric for Explanations}

Many commonly used evaluation metrics for text-based content like BLEU \citep{papineni2002bleu} and ROUGE \citep{lin2004rouge} treat human-annotated answers as the absolute gold standard
without questioning or attempting to evaluate their quality.
One established evaluation metric called \texttt{Simulatability} score derives from Human Simulatability \citep{doshi2017towards} and can examine gold explanations. It simply measures the change in a baseline model prediction performance, 
depending on whether the explanation is provided as the input. Previous works \citep{chandrasekaran2018explanations, yeung2020sequential, hase2020leakage, wiegreffe2020measuring, poursabzi2021manipulating, rajagopal2021selfexplain} have demonstrated the usefulness of \texttt{Simulatability} score for evaluating explanation quality. However, this metric has a couple of inherent disadvantages. First, it only considers the helpfulness of explanations
on a baseline model, where we show that explanations provide different helpfulness during fine-tuning and inference through our experiment in Section~\ref{pe}. In addition, model performance could also differ when we transform 
the original task into other tasks, such as turning a classification task into a multiple-choice task with different input data formats. 

In order to objectively evaluate human-annotated explanations, we define a new evaluation metric based on the \texttt{Simulatability} score that complements both drawbacks of \texttt{Simulatability} by considering the helpfulness of explanations both at fine-tuning and inference with the help of a unified structure to minimize the impact of task differences. Other works~\citep{carton-etal-2020-evaluating} attempted to evaluate and categorize different characteristics of explanations, but many of them~\citep{chan2022frame, deyoung-etal-2020-eraser} still treat human-annotated explanations as the gold standard.

\subsection{Usage of Explanations for SOTA models}
\label{sec:exp_in_models}

Existing works have been exploring circumstances in which explanations could improve model performance; for example, \citet{hase2021can} argues that explanations are most suitable for use as model input for predicting, and \citet{kumar-talukdar-2020-nile} proposed a system to generate label-specific explanations for the NLI task specifically. Some recent works have tried to generate better explanations with a self-rationalization setting \citep{wiegreffe2020measuring, marasovic2021few}, where a model is asked to generate the prediction label and explanation simultaneously. We conduct a preliminary experiment to find the best model setting to leverage explanations in Section~\ref{section_pe1}. 

There exists many recent works \citep{paranjape2021prompting, liu2021generated, chen2022can} that explore the usage of prompts to complete explanations, generate additional information for the original task, or examine whether generated explanations can provide robustness to adversarial attacks. 
\citet{ye2022unreliability} showed that simply plugging explanations into a prompt does not always boost the in-context learning performance, and model-generated explanations can be unreliable for few-shot learning. Another related line of research focuses on extracting or generating explanations with a unified framework \citep{chan2022unirex} or with a teachable reasoning system that generates chains of reasoning \citep{dalvi2022towards}.

%% file: sections/Unified_Structure.tex

\section{Unified Structure}
\label{section_us}

\begin{table}[t]
\centering
\resizebox{.45\textwidth}{!}{%
\small
\begin{tabular}{lccc}
\toprule

\multirow{2}{*}{\textbf{\begin{tabular}[l]{@{}l@{}} Explanations  as\\  Input vs Output \end{tabular}}}  & \multicolumn{3}{c}{\textbf{Fine-tune Setting} }\\
\cmidrule{2-4}
    &\hfil \Baseline &\hfil Self-rationalization &  \Infusion \\

\cmidrule(lr){1-1} \cmidrule(lr){2-2} \cmidrule(lr){3-3} \cmidrule(lr){4-4} 
    
CoS-E v1.0  &   0.695   &   0.646   &   \textbf{0.878}   \\ 
ECQA        &   0.572   &   0.513   &   \textbf{0.989}   \\ 

\bottomrule
\end{tabular}
}

\vspace{-0.5em}

\caption{ Preliminary experiment results of using explanations as part of Input(\Infusion) vs. Output(Self-rationalization) vs. without explanations (\Baseline) on CoS-E and ECQA datasets. }
\vspace{-1em}
\label{tab:pe_1}
\end{table}

While popular metrics like BLEU and ROUGE can evaluate text coherence and similarity, 
one critical aspect of explanations is how beneficial they can be. 
Thus, we want to develop a metric that objectively evaluates explanations' utility towards model performance. Furthermore, we expect that such a metric can systematically demonstrate how good or bad the explanations are; for example, it could objectively measure what `noisy' means in a human study (e.g., from previous works on CoS-E). 

With the advantage of sequence-to-sequence models like T5 that can map different types of language tasks into generation tasks, we can control and minimize the influence of varying task formats on model performance while evaluating the helpfulness of explanations by leveraging a unified data format. We realize that existing datasets with human-annotated explanations are mostly either multiple-choice tasks or classification tasks. The classification task could be viewed as a multiple-choice task where the labels are indeed choices. Inspired by several previous works that manipulated prompts for sequence-to-sequence models \citep{marasovic2021few, liu2021generated}, we incorporate a few well-defined words as template-based prompts for the unified data structure to indicate the task content and corresponding explanations.

Examples shown in Figure~\ref{fig:unified_format}  explain how we map various tasks into a unified multiple-choice generation task. We propose two settings: no explanations (\Baseline) and explanations as additional input (\Infusion). 
Here we explain how each prompt addresses a different part of the data content: 
1) `\textit{explain:}' is followed by the question content, 2) `\textit{choice-$n$:}' is followed by each candidate answer, and 3) a special token `\textit{<sep>}' separates the explanations from the task content, while the explanations in \Infusion are led by `\textit{because}' so that the model knows that the explanation text explains the task content. For datasets like CoS-E and ECQA, we leverage the original task as the question content. On the other hand, we define fixed question prompts for e-SNLI: ``\textit{what is the relation between [Premise] and [Hypothesis]?}'', and for ComVE: ``\textit{which sentence is against commonsense?}'' to specify corresponding tasks to models.

%% file: sections/Preliminary_Experiment.tex
\section{Preliminary Experiment}
\label{pe}
\subsection{Utilizing Explanations as Part of Input vs Part of Output}
\label{section_pe1}

As described in Section~\ref{sec:exp_in_models}, recent works have been exploring various circumstances that human-annotated explanations could help in different aspects.
We hypothesize that leveraging explanations as additional input with the original task input allows models to use explanations for better prediction, while the self-rationalization~\cite{marasovic2021few} setting, which generate explanations along with labels, complicates the prediction task for the models and may lead to a performance decrease. In addition, the generated explanations from self-rationalization systems are not explicitly being used for label prediction. To justify our hypothesis, we conduct a preliminary experiment on CoS-E v1.0 and ECQA datasets. 

We fine-tune three T5-base models on each dataset with three different settings: \Baseline, \Infusion, and explanations as additional output (\textit{Self-Rationalization }hereinafter). For each model, we maintain the same setting during fine-tuning and inference. For example, the model fine-tuned with \Infusion will also take data under \Infusion during inference. We leverage the unified structure for \Baseline and \Infusion shown in Figure~\ref{fig:unified_format} and make minor adjustments for the self-rationalization setting accordingly (shown in Appendix~\ref{app:self-rationalization}). 

The experiment results are shown in Table~\ref{tab:pe_1}. We notice that the self-rationalization setting performs worse than the \Baseline, which is aligned with our assumption. On the other hand, the \Infusion setting surprisingly achieves significant improvement on CoS-E, which was considered `noisy' by previous works, demonstrating that the CoS-E explanations are indeed helpful toward models. The \Infusion setting also approaches nearly complete correctness on the ECQA dataset.

\begin{figure}[!t]
    \centering
    \subfloat[CoS-E v1.0]{%
        \includegraphics[clip,width=.95\columnwidth]{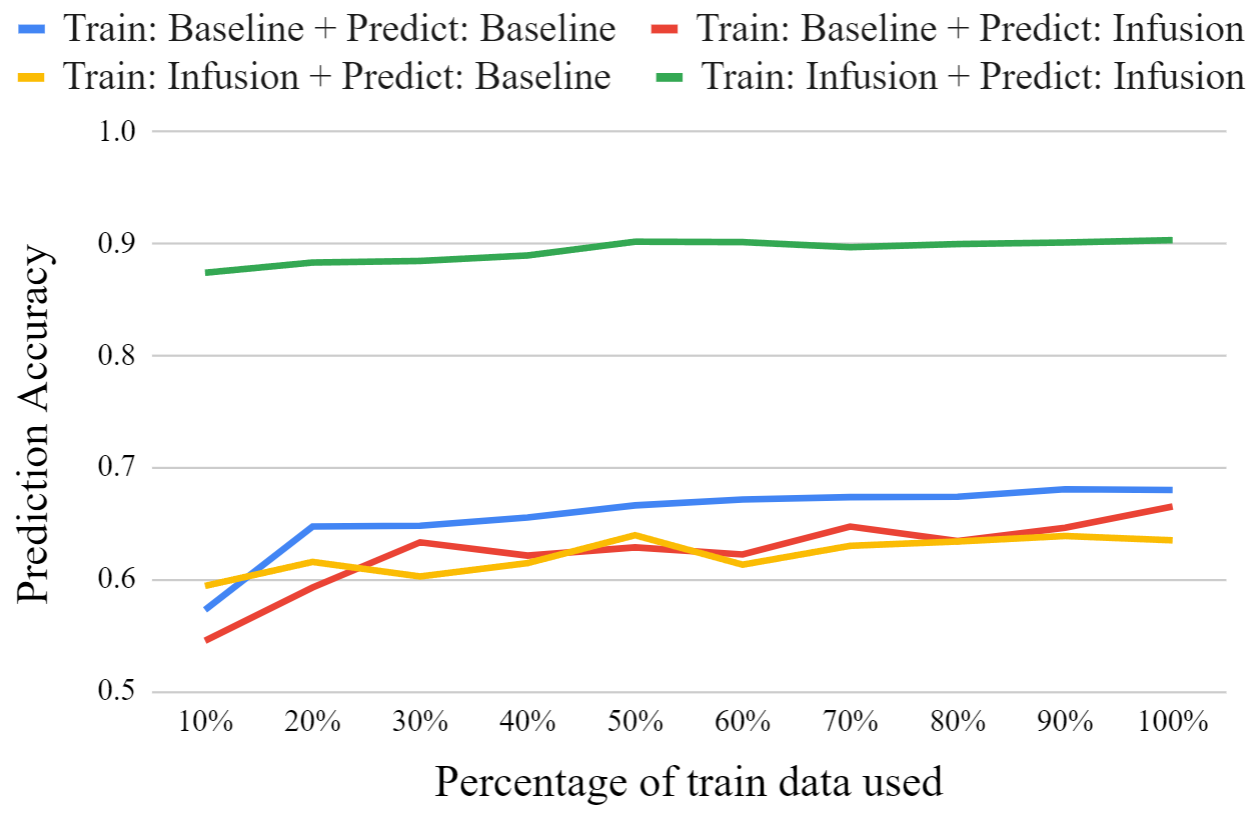}%
    }
    
    \subfloat[ECQA]{%
        \includegraphics[clip,width=.95\columnwidth]{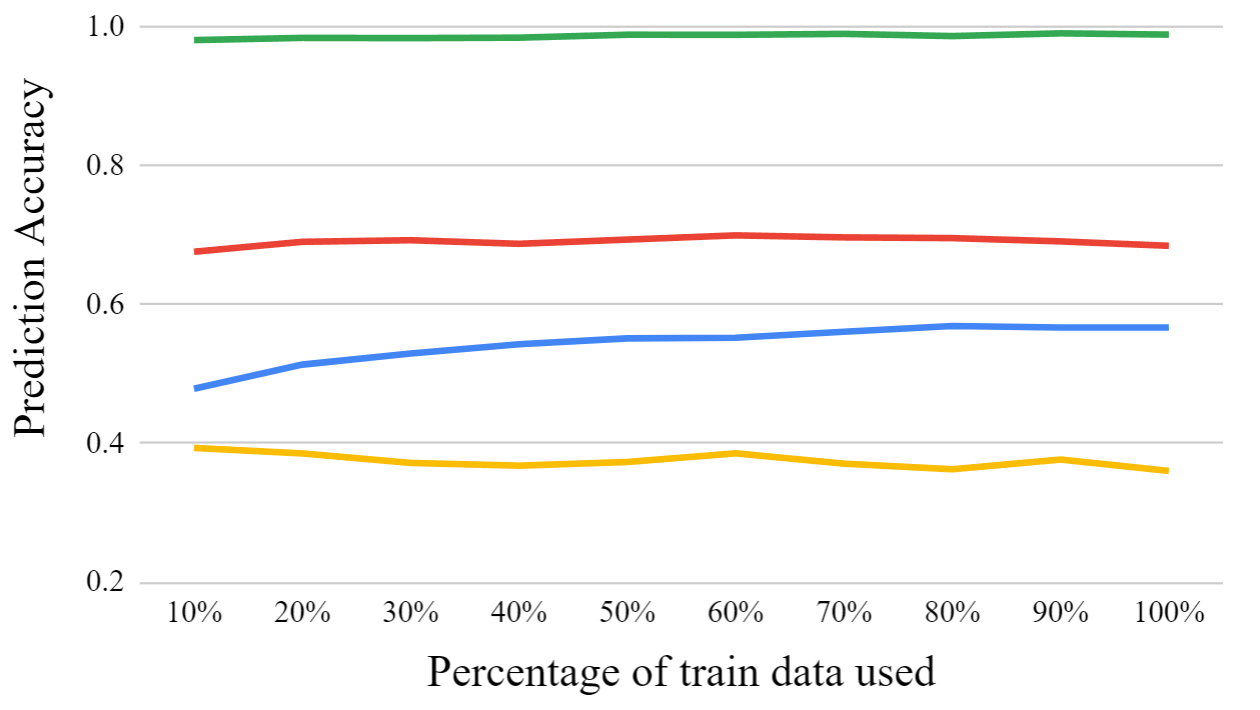}%
    }
    \vspace{-0.5em}
    \caption{ Explanations as partial input on CoS-E v1.0 (top) and ECQA (bottom) with different amounts of training data. We perform fine-tuning and predicting for both \Baseline and \Infusion settings. }
    
    \vspace{-1em}
    \label{fig:pe2_cose}
\end{figure}

\subsection{Explanations as Partial Input During Fine-Tuning}

To examine the utility of explanations to the models during fine-tuning, we perform an in-depth experiment with the \Baseline and \Infusion setting
while varying the amounts of training data used for fine-tuning. 
First, 
we randomly 
select nine sub-datasets with amounts of data ranging from 10\% to 90\% of the training data in each dataset used in the first preliminary experiment. Then, for each sub-dataset, we fine-tune three models with different random seeds for sampling and fine-tuning, then acquire the averaged prediction performance. As a result, for each CoS-E v1.0 and ECQA dataset, we get 60 models fine-tuned with varying amounts of data for both the \Baseline and \Infusion setting, including the models fine-tuned on full training data, then perform prediction with the \Baseline and \Infusion settings. We maintain the same hyper-parameters across the models fine-tuned for this experiment and report them in Appendix~\ref{app:hyper_pe}.

The two diagrams in Figure~\ref{fig:pe2_cose} show the experiment results on two datasets (detailed results in Table~\ref{tab:app_pe_results} in the appendix). Different colors denote different fine-tuning and inference settings. We conclude with a few interesting observations:\\
\indent 1. By looking at \textbf{\textcolor{cYELLOW}{yellow}} (model fine-tuned with \Infusion and predict with \Baseline) and \textbf{\textcolor{cGREEN}{green}} (model fine-tuned and predict with \Infusion) line, we notice adding more training data during fine-tuning does not significantly improve model performance, suggesting that \textbf{the fine-tuning process is not teaching the model with new knowledge that is 
conveyed in the explanations}. \\
\indent 2. By comparing \textbf{\textcolor{cYELLOW}{yellow}} and \textbf{\textcolor{cBLUE}{blue}} (model fine-tuned and predict with \Baseline) line in each diagram, we notice the models fine-tuned with \Infusion perform worse than baseline models without explanations during inference, demonstrating that \textbf{fine-tuning with \Infusion teaches the models to rely on the explanations to predict}. \\
\indent 3. By comparing \textbf{\textcolor{cRED}{red}} (model fine-tuned with \Baseline and predict with \Infusion) and \textbf{\textcolor{cBLUE}{blue}} line in each diagram, we observe the baseline models for CoS-E perform worse while predicting with explanations. In contrast, the baseline models for ECQA consistently exceed baseline performance significantly, which demonstrates that \textbf{the helpfulness of explanations on baseline models in CoS-E is much worse than the ones in ECQA}, which is aligned with some previous works. \\
\indent 4. By comparing \textbf{\textcolor{cGREEN}{green}} and \textbf{\textcolor{cBLUE}{blue}} lines in both diagrams, we notice that explanations in CoS-E can contribute to substantial improvement during inference on models fine-tuned with \Infusion setting. This observation shows that \textbf{explanations in CoS-E are able to provide helpfulness to models during fine-tuning, even though they were considered `noisy' by humans in previous works.} \\
\indent 5. By comparing \textbf{\textcolor{cRED}{red}} and \textbf{\textcolor{cGREEN}{green}} lines in both diagrams, we can observe that in order to take full advantage of explanations, \textbf{it is beneficial to fine-tune a model even with a small amount of data that incorporates the explanations.} Such fine-tuning can lead to a substantial improvement.


\begin{figure}[t]

\vspace{-1.5em}
\begin{align*} 
\metricname = & \big(Accu(M_{\Infusion}^{\Infusion}) - Accu(M_{\Baseline}^{\Baseline}) ) \\ 
            + & \big(Accu(M_{\Baseline}^{\Infusion}) - Accu(M_{\Baseline}^{\Baseline}) ) 
\end{align*}
\vspace{-1.5em}

\caption{ The formula of our \metricname metric. $M$ denotes a model and the subscript/superscript denotes $M_{finetune \, setting}^{predict \, setting}$. The \texttt{Simulatability} score only considers the second part within our formula. }
\label{fig:formula}
\vspace{-1.5em}
\end{figure}

This experiment shows that explanations provide different degrees of utility during fine-tuning and inference. Thus, we should consider both situations while evaluating the helpfulness of explanations.


%% file: sections/eval.tex
\section{Our Metric and Evaluation}

\subsection{Our \metricname Metric}

Based on our observations from the preliminary experiments, 
we propose a novel evaluation metric that extends the \texttt{Simulatability} score.
Figure~\ref{fig:formula} shows the formula of our \metricname metric: it evaluates the helpfulness of explanations with the sum of two parts: 
at fine-tuning, where two models are fine-tuned with \Baseline and \Infusion settings correspondingly, we calculate the prediction accuracy difference using the same data format that was used during fine-tuning for each model; and at inference, we fine-tune only one model with \Baseline setting and calculate the prediction accuracy difference between \Infusion and \Baseline settings. 

The second part of our metric is indeed the \texttt{Simulatability} metric. 
We observe that fine-tuning a model with data that incorporates explanations can provide substantial benefits. However, the \texttt{Simulatability} score fails to account for this component and only considers the model performance improvement that uses explanations at inference without fine-tuning first.
For the models fine-tuned with \Baseline setting, we believe pre-trained SOTA large-scale models have the ability to understand the additional content at the input to a certain extent. The addition of explanations at input during inference will show whether it can provide helpfulness to a baseline model without additional supervision, while the models fine-tuned with \Infusion setting will rely more on the explanation part of the input for inference. 

A positive score demonstrates that the explanations can provide overall helpfulness for better prediction, while a negative score does not necessarily mean the explanations are not helpful. Instead, a negative score indicates that the explanations lead to the model's performance drop in at least one part of the evaluation. Researchers can further analyze the intermediate score for each part.
As a result, the score ranges theoretically from -2 to 2. 

\begin{table*}[t]
\centering
\resizebox{.95\textwidth}{!}{%

\begin{tabular}{lccccc} 
\toprule


\multirow{2}{*}{ \textbf{T5-base} }     
&   \multirow{2}{*}{ $M_{finetune + \Baseline}^{predict + \Baseline}$ }   
&   \multirow{2}{*}{ $M_{finetune + \Baseline}^{predict + \Infusion}$ }    
&   \multirow{2}{*}{ \begin{tabular}{@{}c@{}} \textbf{\texttt{Simulatability}} \\ \textbf{Score} \end{tabular} }
&   \multirow{2}{*}{ $M_{finetune + \Infusion}^{predict + \Infusion}$ }    
&   \multirow{2}{*}{ \begin{tabular}{@{}c@{}} \textbf{\metricname} \\ \textbf{Score} \end{tabular}  }  \\                                   
&   &   &   &   &   \\

\cmidrule(lr){1-1} \cmidrule(lr){2-2} \cmidrule(lr){3-3} \cmidrule(lr){4-4} \cmidrule(lr){5-5} \cmidrule(lr){6-6} 

ECQA          &   $0.572$   &   $0.746$   &   \textbf{0.174}   &   $0.989$   &    \textbf{0.591}  \\
CoS-E v1.11   &   $0.608$   &   $0.610$   &   \textbf{0.002}   &   $0.803$   &   \textbf{0.197}  \\
CoS-E v1.0    &   $0.695$   &   $0.645$   &   \textbf{-0.05}   &   $0.878$   &   \textbf{0.133}  \\
e-SNLI        &   $0.907$   &  $0.676$    &  \textbf{-0.231}   &   $0.981$   &   \textbf{-0.157}  \\
ComVE         &   $0.88$    &   $0.527$   &  \textbf{-0.353}   &   $0.949$   &    \textbf{-0.284}   \\

\bottomrule
\end{tabular}
}

\medskip

\resizebox{.95\textwidth}{!}{%

\begin{tabular}{lccccc} 
\toprule




\multirow{2}{*}{ \textbf{BART-base}  }     
&   \multirow{2}{*}{ $M_{finetune + \Baseline}^{predict + \Baseline}$  }                 
&   \multirow{2}{*}{ $M_{finetune + \Baseline}^{predict + \Infusion}$  }  
&   \multirow{2}{*}{ \begin{tabular}{@{}c@{}}  \textbf{\texttt{Simulatability} } \\ \textbf{Score }\end{tabular} }
&   \multirow{2}{*}{ $M_{finetune + \Infusion}^{predict + \Infusion}$  }  
&   \multirow{2}{*}{ \begin{tabular}{@{}c@{}} \textbf{\metricname} \\ \textbf{Score} \end{tabular}  } \\  

&   &   &   &   &   \\

\cmidrule(lr){1-1} \cmidrule(lr){2-2} \cmidrule(lr){3-3} \cmidrule(lr){4-4} \cmidrule(lr){5-5} \cmidrule(lr){6-6} 




ECQA          &   0.428   &   0.438   &   \textbf{0.010}   &   0.901   &   \textbf{0.483}  \\
CoS-E v1.11   &   0.443   &   0.449   &   \textbf{0.006}   &   0.700   &   \textbf{0.263}  \\
CoS-E v1.0    &   0.512   &   0.486   &   \textbf{-0.026}   &   0.790   &   \textbf{0.252}  \\

e-SNLI        &  0.888     &   0.658   &  \textbf{-0.23}   &  0.978   &    \textbf{-0.14}  \\
ComVE         &   0.812    &   0.596   &  \textbf{-0.216}   &   0.864   &    \textbf{-0.164}   \\
\bottomrule
\end{tabular}
}

\caption {Evaluation results of human-annotated explanations in 5 datasets with our \metricname score and \texttt{Simulatability} score.
The tables above and below correspond to models fine-tuned on T5-base and BART-base, respectively. The \texttt{Simulatability} score only considers $M_{finetune + \Baseline}^{predict + \Baseline}$ and $M_{finetune + \Baseline}^{predict + \Infusion}$, while our \metricname score considers $M_{finetune + \Infusion}^{predict + \Infusion}$ additionally.}
\vspace{-1em}
\label{tab:eval_results}
\end{table*}

\subsection{Evaluation}

We evaluate human-annotated natural language explanations across five popular datasets using our \metricname metric and the \texttt{Simulatability} score. To justify that our metric is less biased by different model architectures and to examine the influence of models fine-tuned with different settings towards the prediction performance, we perform experiments on both T5 and BART models.
The proposed unified data format is applied to the experiments for our metric and the \texttt{Simulatability} score to make it a more robust baseline. 

We maintain the same fine-tuning hyper-parameters for all the experiments (details in Appendix~\ref{app:hyper_eval}). The only exception is for the e-SNLI dataset, which has about 10x the size (549,367 data instances) of training data compared to the other datasets. Therefore, we only fine-tune models on the e-SNLI dataset with two epochs. Furthermore, we leverage the special token `\textit{<s>}' for BART that was already used during the pre-training process instead of using and adding the special token `\textit{<sep>}' to BART tokenizer during fine-tuning. We present the evaluation results in Table~\ref{tab:eval_results}.


\subsection{Findings}


\paragraph{Our results justify the intuition that human-annotated explanations can still provide benefits toward model prediction, even if they were evaluated as low-quality by humans in prior literature.}
By first comparing the models' prediction results over two architectures, the result shows all models fine-tuned on T5-base outperform those fine-tuned on BART-base with the same setting, mainly with a significant margin.

Despite apparent performance differences between model architectures, by looking at the orderings of datasets in both tables, which are based on our \metricname score, We can easily observe that \metricname score provides the same ranking result for the quality of explanations in 5 datasets over two model architectures. 
Our \metricname score (Table~\ref{tab:eval_results}) ranks the explanation quality of the five datasets in the following order regardless of model architectures:



\vspace{-1.5em}
\begin{align*}
\small \textbf{ECQA > CoS-E v1.11 > CoS-E v1.0 > e-SNLI > ComVE}  
\end{align*}

According to the \metricname score, explanations in ECQA have the best quality among the five datasets. Especially, explanations in ECQA are much better than the ones in both CoS-E datasets, which is consistent with previous works' consensus. It is worth noticing that both CoS-E datasets achieve positive \metricname scores, though significantly lower than the ones for ECQA, demonstrating that explanations in CoS-E datasets still have positive overall helpfulness for models' prediction performance even though they are considered `low quality and noisy' from human experiments~\citep{sun2022investigating}. 

\paragraph{Our \metricname score can rank explanation quality consistently across all five datasets on two models while the \texttt{Simulatability} falls short.} 
On the other hand, the \texttt{Simulatability} score cannot provide a consistent ranking of explanation quality on the two models. Instead, the \texttt{Simulatability} score provides two distinct rankings:
\vspace{-1em}
\begin{align*}
\small   \textbf{\underline{T5-base}:}&    \\
&\small  \textbf{ECQA >  CoS-E v1.11} > \\
&\small  \textbf{CoS-E v1.0 > \textcolor{red}{ e-SNLI >  ComVE} }   \\
\small   \textbf{\underline{BART-base}:}&  \\
&\small  \textbf{ECQA > CoS-E v1.11} > \\
&\small  \textbf{CoS-E v1.0 > \textcolor{red}{ComVE> e-SNLI} }
\end{align*}
From Table~\ref{tab:eval_results}, the \texttt{Simulatability} score ranks e-SNLI and ComVE reversely on BART compared with T5 models, indicating \texttt{Simulatability} score could be more affected by different model architectures even with the unified data structure.  


One advantage of using our \metricname score to evaluate the quality of explanations is that we can analyze the score by class or intermediate results from fine-tuning or inference. For instance, we observe that the \metricname scores for e-SNLI with T5 and BART models are both negative, indicating that the helpfulness of explanations in e-SNLI could be limited. However, by looking into the intermediate results, though the baseline models perform significantly worse while predicting with \Infusion than with \Baseline setting, the models that are fine-tuned with \Infusion still outperform the baseline models while predicting with \Infusion, justifying the explanations indeed provide improvements under this setting. 
When we further decompose the \metricname score of e-SNLI by category, we acquire \textit{0.13/-0.483/0.094} on T5-base and \textit{0.015/-0.227/-0.271} on BART-base corresponds to \textit{entailment/neutral/contradiction}.


\paragraph{We speculate that the helpfulness of human-annotated explanations to models highly depends on the task (e.g., the `contradiction' label categories) and the explanation format (e.g., counter-factorial styles).} 
We notice that the models fine-tuned on T5 and BART have more than a 40\% prediction accuracy drop on data with `neutral' labels when they are fine-tuned with \Baseline and predicted with \Infusion. 
In addition, we observe that the fine-tuned BART models have about a 40\% prediction accuracy drop on data with ground-truth `contradiction' labels. 
We suspect human annotators behave differently while providing explanations for different categories in e-SNLI. 
For instance, 
humans tend to provide counter-factorial explanations or use negation connotations to explain why two sentences are `neutral' or `contradiction' categories. 
Some representative examples for each class are provided in Appendix~\ref{tab:app_esnli_examples}.
Such behavior's tendency to use negation connotations in explanations for specific categories may increase the difficulty for the models to interpret the information and lead to false predictions eventually. 

From Table~\ref{tab:eval_results}, ComVE ranks worst among the five datasets in both tables, indicating the explanations in ComVE are the least helpful for the models to either fine-tune or predict with. Since the ComVE task asks models to predict which sentence is more likely \textbf{\emph{against}} commonsense, the question itself implies a negation connotation. Likewise, many ComVE explanations contain negation, such as the one in Figure~\ref{fig:unified_format}. The concept of negation has always been a complex concept for machines. Although both T5 and BART models fine-tuned with the \Baseline setting can perform relatively well on ComVE, the addition of explanations that largely contain negation during inference is likely to create more difficulties for the models to understand and eventually lead to false prediction.

\paragraph{Our hypothesis on counter-examples or negation annotations in human-annotated explanations can find support from many recent works.} A recent analysis~\citep{Joshi2022AreAS} claimed that negation connotations have high necessity but low sufficiency to describe the relation between features and labels.
In addition, counterfactually-augmented data may prevent models from learning unperturbed robust features and exacerbate spurious correlations~\citep{joshi2021investigation}. Therefore, we suggest human annotators avoid using counter-examples while providing explanations. Instead, using precise words to describe the degree of relations between concepts will be preferable and provide better helpfulness to models. 

Nevertheless, these models can correctly understand explanations for all categories after being fine-tuned with the \Infusion setting. Worth pointing out that ECQA explanations are summarized from positive and negative properties for each candidate choice which also contains negation words. However, those negation words mostly appear in negative properties for wrong choices. As a result, we notice the pre-trained baseline models can leverage ECQA explanations with \Infusion during the predicting process and achieve performance improvement. Since we are the first to discover such a class-level drop on e-SNLI by using \metricname score, we only propose our hypothetical assumption and leave a definitive study for future work.  


%% file: sections/conclusion.tex
\section{Conclusion}

In this paper, we objectively evaluate human-annotated natural language explanations from the perspective of measuring their helpfulness towards models' prediction.  
We conduct two preliminary experiments and 
based on the findings from the preliminary study, we define an evaluation metric that considers the explanations' helpfulness at both fine-tuning and inference stages;
We also propose a unified prompt-based data format that minimizes the influence of task differences by mapping various tasks into a unified multiple-choice generation task.
Our experiment with human-annotated explanations in five popular large-scale datasets over two sequence-to-sequence model architectures demonstrates that our metric can consistently reflect the relative ranking of explanation qualities among five datasets while the \texttt{Simulatability} score falls short. 
Our work lays a stepstone towards a high-quality human-AI collaboration future for data annotation job~\cite{wang2019human}, and we recommend  
researchers perform similar quality checks while collecting human-annotated explanations in the future.



%% file: sections/limitations.tex
\section{Limitations}

In this paper, we evaluate the quality of human-annotated natural language explanations towards the models' prediction performance on multiple datasets. Although it is a natural step that our evaluation metric could be generalized to evaluate the helpfulness of model-generated explanations, we would like to caution that:
our metric and evaluation experiment requires the models to generate explanations for the train split data, then use the data with generated explanations to fine-tune the second model with the \Infusion setting, which may not be suitable for those systems that are trained on train split data. 
In addition, we acknowledge that the human-annotated explanations are very expensive to collect, thus, a better mechanism (e.g., Active-Learning approaches~\cite{yao2023active}) is needed to improve human annotators' performance.

%% file: sections/ethics.tex
\section{Ethics Statement}
We do not see potential ethical concerns or misuse of the proposed evaluation method. One potential risk, though minimal, could be the misinterpretation of the findings of this paper. We would like to caution readers that a higher score of our metric may not necessarily reflect a higher quality perceived by humans, as the evaluation metric only measures the explanation's benefit from the modeling perspective, and it is only one of the many possible ways of automatically evaluating the quality of natural language explanations.

%% file: sections/ack.tex
\section*{Acknowledgements}

This work was supported by the Rensselaer-IBM AI Research Collaboration (http://airc.rpi.edu), part of the IBM AI Horizons Network (http://ibm.biz/AIHorizons).

%% file: sections/appendix.tex
\section*{Appendix}

\section{Implementation of self-rationalization format}
\label{app:self-rationalization}

\begin{figure*}[t]
    \centering
    \includegraphics[width=.95\textwidth]{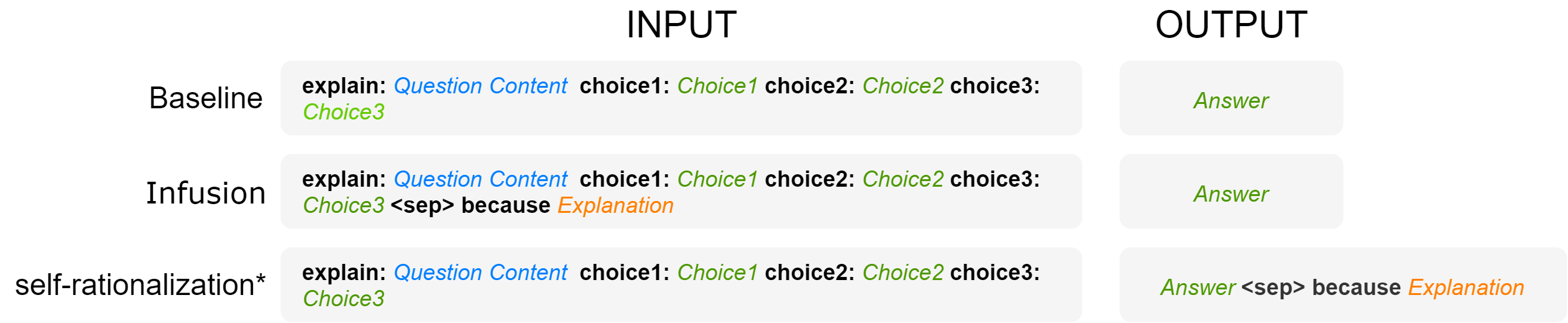}
    \caption{ The unified structure of \Baseline, \Infusion, and self-rationalization settings. Bold text are fixed prompts for each dataset. }
    \vspace{-1em}
    \label{fig:appendix_format}
\end{figure*}

We show the implementation of the self-rationalization setting proposed by \citet{marasovic2021few} and put it together in Figure~\ref{fig:appendix_format} with our proposed unified structure of the \Baseline and \Infusion setting. 

\section{Experiment Hyper-Parameters}

We perform all the computational experiments on a Google Colab instance with a single Nvidia V100 GPU and 50 Gigabytes of RAM.

\subsection{Hyper-parameter for Preliminary Experiment}
\label{app:hyper_pe}
For the preliminary experiment of utilizing explanations as part of input V.S. part of the output, we leverage the following hyper-parameters for all models with different data structures: 
$max\_len: 512$, $target\_max\_len: 64$, $train\_batch\_size: 1$, $learning\_rate: 5e^{-5}$, $num\_train\_epochs: 12$. 

For the preliminary experiment of explanations as partial input during fine-tuning, we maintain the following hyper-parameters for all models fine-tuned with partial/full train data of CoS-E and ECQA datasets: $max\_len: 512$, $target\_max\_len: 16$, $train\_batch\_size: 1$, $learning\_rate: 1e^{-4}$, $num\_train\_epochs: 6$. 

\subsection{Hyper-parameter for Explanation Evaluation with five Datasets}
\label{app:hyper_eval}
For the evaluation of human-annotated explanations on 5 different datasets, we maintain the following hyper-parameters for all the models: $max\_len: 512$, $target\_max\_len: 64$, $train\_batch\_size: 1$, $learning\_rate: 5e^{-5}$, $num\_train\_epochs: 12$. The only exception is the e-SNLI dataset, which has about 10x the size (549,367 data instances) of training data compared to the other datasets. Therefore, we only fine-tune models on the e-SNLI dataset with two epochs.

\section{Results for Preliminary Experiment - Explanations as Partial Input During Fine-tuning}

We randomly shuffle three seeds to select the subset of data and fine-tune the model for the preliminary experiment of explanations as partial input during fine-tuning. The detailed results of each experiment and average accuracy are reported in Table~\ref{tab:app_pe_results}.

\begin{table*}[t]
\centering

\resizebox{.9\textwidth}{!}{%
\begin{tabular}{lcccccccccc}
\toprule
\multicolumn{11}{c}{\textbf{Fine-tune with \Baseline on CoS-E v1.0} } \\
\cmidrule(lr){2-11}
        &   10\%    &   20\%    &   30\%    &   40\%    &   50\%    &   60\%    &   70\%    &   80\%    &   90\%    &   1   \\

\cmidrule(lr){1-1} \cmidrule(lr){2-2} \cmidrule(lr){3-3} \cmidrule(lr){4-4} \cmidrule(lr){5-5} \cmidrule(lr){6-6} \cmidrule(lr){7-7} \cmidrule(lr){8-8} \cmidrule(lr){9-9} \cmidrule(lr){10-10} \cmidrule(lr){11-11}

\multirow{3}{*}{\centering \begin{tabular}[c]{@{}c@{}} Predict \\ \Baseline \end{tabular} }  &  0.583   &	0.656   &	0.638   &	0.658   &	0.661   &	0.670   &	0.674   &	0.678   &	0.697   &	0.676   \\
        &   0.550   &	0.644   &	0.664   &	0.650   &	0.666   &	0.667   &	0.667   &	0.682   &	0.668   &	0.682    \\
        &   0.584   &	0.64    &	0.64    &	0.655   &	0.670   &	0.675   &	0.677   &	0.66    &	0.674   &	0.68    \\
\midrule  
Average &   0.572    &	0.647    &	0.647    &	0.655    &	0.665    &	0.671    &	0.673    &	0.673   &	0.680    &	0.679    \\
\midrule 

\multirow{3}{*}{\centering \begin{tabular}[c]{@{}c@{}} Predict \\ \Infusion \end{tabular} }  &  0.586   &	0.586   &	0.625   &	0.633   &	0.596   &	0.621   &	0.663   &	0.655   &	0.649   &	0.676   \\
        &   0.561   &	0.591   &	0.642   &	0.609   &	0.656   &	0.630   &	0.618   &	0.650   &	0.641   &	0.652   \\
        &   0.525   &	0.6     &	0.631   &	0.62    &	0.631   &	0.614   &	0.658   &	0.595   &	0.647   &	0.665   \\
\midrule
Average &   0.545   &	0.592   &	0.632   &	0.621   &	0.628   &	0.622   &	0.647   &	0.634   &	0.645   &	0.664   \\
\bottomrule
\end{tabular}
}

\medskip

\resizebox{.9\textwidth}{!}{%
\begin{tabular}{lcccccccccc} 
\toprule
\multicolumn{11}{c}{\textbf{Fine-tune with \Infusion on CoS-E v1.0} } \\
\cmidrule(lr){2-11}
        &   10\%    &   20\%    &   30\%    &   40\%    &   50\%    &   60\%    &   70\%    &   80\%    &   90\%    &   1   \\

\cmidrule(lr){1-1} \cmidrule(lr){2-2} \cmidrule(lr){3-3} \cmidrule(lr){4-4} \cmidrule(lr){5-5} \cmidrule(lr){6-6} \cmidrule(lr){7-7} \cmidrule(lr){8-8} \cmidrule(lr){9-9} \cmidrule(lr){10-10} \cmidrule(lr){11-11}

\multirow{3}{*}{\centering \begin{tabular}[c]{@{}c@{}} Predict \\ \Baseline \end{tabular} }  &  0.588   &	0.622   &	0.617   &	0.613   &	0.635   &	0.616   &	0.615   &	0.625   &	0.652   &	0.629   \\
        &   0.592   &	0.614   &	0.573   &	0.610   &	0.650   &	0.592   &	0.632   &	0.64    &	0.610   &	0.64    \\
        &   0.601   &	0.609   &	0.615   &	0.618   &	0.631   &	0.629   &	0.641   &	0.635   &	0.652   &	0.634   \\
\midrule

Average &   0.594   &	0.615   &	0.602   &	0.614   &	0.639   &	0.612   &	0.629   &	0.633   &	0.638   &	0.634   \\
\midrule 
\multirow{3}{*}{\centering \begin{tabular}[c]{@{}c@{}} Predict \\ \Infusion \end{tabular} }  &  0.867   &	0.874   &	0.884   &	0.889   &	0.902   &	0.894   &	0.890   &	0.886   &	0.910   &	0.904    \\
        &   0.875   &	0.888   &	0.881   &	0.890   &	0.898   &	0.901   &	0.9     &	0.901   &	0.896   &	0.895    \\
        &   0.877   &	0.885   &	0.887   &	0.887   &	0.903   &	0.907   &	0.898   &	0.910   &	0.894   &	0.908    \\
\midrule

Average &   0.873   &	0.882   &	0.884   &	0.889   &	0.901   &	0.901   &	0.896   &	0.899   &	0.900   &	0.902    \\
\bottomrule
\end{tabular}
}

\medskip

\resizebox{.9\textwidth}{!}{%
\begin{tabular}{lcccccccccc}
\toprule
\multicolumn{11}{c}{\textbf{Fine-tune with \Baseline on ECQA} } \\
\cmidrule(lr){2-11}
        &   10\%    &   20\%    &   30\%    &   40\%    &   50\%    &   60\%    &   70\%    &   80\%    &   90\%    &   1   \\

\cmidrule(lr){1-1} \cmidrule(lr){2-2} \cmidrule(lr){3-3} \cmidrule(lr){4-4} \cmidrule(lr){5-5} \cmidrule(lr){6-6} \cmidrule(lr){7-7} \cmidrule(lr){8-8} \cmidrule(lr){9-9} \cmidrule(lr){10-10} \cmidrule(lr){11-11}

\multirow{3}{*}{\centering \begin{tabular}[c]{@{}c@{}} Predict \\ \Baseline \end{tabular} }  &  0.495   &	0.522   &	0.528   &	0.553   &	0.550   &	0.550   &	0.554   &	0.569   &	0.561   &	0.562   \\
        &   0.471   &	0.505   &	0.525   &	0.533   &	0.549   &	0.561   &	0.558   &	0.572   &	0.572   &	0.572   \\
        &   0.469   &	0.511   &	0.533   &	0.541   &	0.553   &	0.545   &	0.569   &	0.564   &	0.566   &	0.565   \\
\midrule  
Average &   0.478   &	0.513   &	0.529   &	0.542   &	0.551   &	0.552   &	0.560   &	0.568   &	0.566   &	0.566   \\
\midrule 
\multirow{3}{*}{\centering \begin{tabular}[c]{@{}c@{}} Predict \\ \Infusion \end{tabular} }  &  0.664   &	0.672   &	0.710   &	0.716   &	0.692   &	0.702   &	0.708   &	0.722   &	0.684   &	0.701   \\
        &   0.685   &	0.682   &	0.673   &	0.697   &	0.681   &	0.682   &	0.694   &	0.677   &	0.699   &	0.641   \\
        &   0.678   &	0.715   &	0.693   &	0.648   &	0.706   &	0.713   &	0.686   &	0.685   &	0.688   &	0.711   \\
\midrule
Average &   0.675   &	0.690   &	0.692   &	0.687   &	0.693   &	0.699   &	0.696   &	0.695   &	0.690   &	0.684   \\
\bottomrule
\end{tabular}
}

\medskip

\resizebox{.9\textwidth}{!}{%
\begin{tabular}{lcccccccccc}
\toprule
\multicolumn{11}{c}{\textbf{Fine-tune with \Infusion on ECQA} } \\
\cmidrule(lr){2-11}
        &   10\%    &   20\%    &   30\%    &   40\%    &   50\%    &   60\%    &   70\%    &   80\%    &   90\%    &   1   \\

\cmidrule(lr){1-1} \cmidrule(lr){2-2} \cmidrule(lr){3-3} \cmidrule(lr){4-4} \cmidrule(lr){5-5} \cmidrule(lr){6-6} \cmidrule(lr){7-7} \cmidrule(lr){8-8} \cmidrule(lr){9-9} \cmidrule(lr){10-10} \cmidrule(lr){11-11}

\multirow{3}{*}{\centering \begin{tabular}[c]{@{}c@{}} Predict \\ \Baseline \end{tabular} }    &   0.417 &   0.406   &   0.402   &   0.395   &   0.381 &   0.379   &   0.365   &   0.379   &   0.375   &   0.374   \\
        &   0.381   &	0.363   &	0.367   &	0.366   &	0.368   &	0.400   &	0.385   &	0.349   &	0.368   &	0.371  \\
        &   0.381   &	0.386   &	0.345   &	0.341   &	0.369   &	0.376   &	0.361   &	0.359   &	0.386   &	0.334  \\
\midrule
Average &   0.393   &	0.385   &	0.371   &	0.367   &	0.373   &	0.385   &	0.370   &	0.362   &	0.376   &	0.360  \\
\midrule 
\multirow{3}{*}{\centering \begin{tabular}[c]{@{}c@{}} Predict \\ \Infusion \end{tabular} }    &   0.974 &	0.983   &	0.983   &	0.989   &	0.985   &	0.988   &	0.989   &	0.984   &	0.990   &	0.992   \\
        &   0.984   &	0.985   &	0.983   &	0.981   &	0.990   &	0.989   &	0.991   &	0.985   &	0.990   &	0.983  \\
        &   0.984   &	0.982   &	0.984   &	0.981   &	0.989   &	0.987   &	0.988   &	0.989   &	0.989   &	0.989  \\
\midrule
Average &   0.980   &	0.983   &	0.983   &	0.984   &	0.988   &	0.988   &	0.989   &	0.986   &	0.990   &	0.988  \\
\bottomrule
\end{tabular}
}

\medskip
\vspace{-10pt}
\caption { Detailed results for the preliminary experiment of explanations as partial input during fine-tuning. }
\vspace{-1em}
\label{tab:app_pe_results}
\end{table*}



\section{Examples of different explanations for each category in e-SNLI dataset}

From our evaluation results, we suspect human annotators behave differently while explaining data with various categories in e-SNLI. For instance, human annotators may explain why two sentences are `entailment' by describing the shared information or similarities conveyed by both sentences, which is easy for models to understand. However, humans tend to provide counter-examples or negations to explain why two sentences are unrelated (neutral) or contradictory rather than explaining their reasoning in a positive way. In Table~\ref{tab:app_esnli_examples}, we show representative examples of data with corresponding explanations for each class.

\begin{table*}[!t]
\centering
\resizebox{.98\textwidth}{!}{%

\begin{tabular}{llll} 

\toprule

Category     &    Premise        &   Hypothesis    &    Explanation  \\

\cmidrule(lr){1-1} \cmidrule(lr){2-2} \cmidrule(lr){3-3} \cmidrule(lr){4-4} 

entailment      &   \begin{tabular}[l]{@{}l@{}} A young family enjoys feeling \\ ocean waves lap at their feet. \end{tabular} 
& A family is at the beach.
& Ocean waves implies the beach. \\

\cmidrule(lr){2-2} \cmidrule(lr){3-3} \cmidrule(lr){4-4} 

&   \begin{tabular}[l]{@{}l@{}} An old man with a package poses \\ in front of an advertisement. \end{tabular}    &   A man poses in front of an ad.    &   \begin{tabular}[l]{@{}l@{}} The word " ad " is short for the word \\ " advertisement ". \end{tabular}    \\

\cmidrule(lr){2-2} \cmidrule(lr){3-3} \cmidrule(lr){4-4} 

&   \begin{tabular}[l]{@{}l@{}} A man reads the paper in a bar \\ with green lighting.  \end{tabular}  &   The man is inside. &    In a bar means the man could be inside.   \\

\midrule

neutral         &   \begin{tabular}[l]{@{}l@{}} An old man with a package poses \\ in front of  an advertisement. \end{tabular} 
& \begin{tabular}[l]{@{}l@{}} A man poses in front of \\ an ad for beer. \end{tabular}  
& Not all advertisements are ad for beer.   \\

\cmidrule(lr){2-2} \cmidrule(lr){3-3} \cmidrule(lr){4-4} 

&   \begin{tabular}[l]{@{}l@{}} A woman with a green headscarf, \\ blue shirt and a very big grin. \end{tabular}  &   The woman is young. &   \begin{tabular}[l]{@{}l@{}} the woman could've been old rather \\ than young \end{tabular} \\

\cmidrule(lr){2-2} \cmidrule(lr){3-3} \cmidrule(lr){4-4} 

&   \begin{tabular}[l]{@{}l@{}} A man reads the paper in a bar \\ with green lighting. \end{tabular}   &   The man is reading the sportspage.  & \begin{tabular}[l]{@{}l@{}} The man could be reading something \\ other than the sportspage. \end{tabular} \\

\midrule

contradiction   &   \begin{tabular}[l]{@{}l@{}} A woman with a green headscarf, \\ blue  shirt and a very big grin. \end{tabular}  
& The woman has been shot. 
& \begin{tabular}[l]{@{}l@{}} There can be either a woman with a very \\ big grin or a woman who has been shot. \end{tabular}   \\

 \cmidrule(lr){2-2} \cmidrule(lr){3-3} \cmidrule(lr){4-4} 

&   \begin{tabular}[l]{@{}l@{}} A man playing an electric guitar \\ on stage. \end{tabular}  &   A man playing banjo on the floor. &   \begin{tabular}[l]{@{}l@{}} The man can't play on stage if he is \\ on the floor.  \end{tabular}\\

 \cmidrule(lr){2-2} \cmidrule(lr){3-3} \cmidrule(lr){4-4} 

&   \begin{tabular}[l]{@{}l@{}} A couple walk hand in hand \\ down a street. \end{tabular}   &   A couple is sitting on a bench.   &   \begin{tabular}[l]{@{}l@{}} The couple cannot be walking and \\ sitting a the same time. \end{tabular} \\

\bottomrule
\end{tabular}
}

\medskip
\vspace{-10pt}
\caption { Representative examples of data with corresponding explanations for each class in e-SNLI.     }
\vspace{-1em}
\label{tab:app_esnli_examples}
\end{table*}